\documentclass[runningheads]{llncs}

\usepackage[T1]{fontenc}
\usepackage{graphicx}
\usepackage{booktabs}
\usepackage{multirow}
\usepackage{array}
\usepackage{xcolor}
\usepackage{url}
\usepackage{caption}
\begin{document}

\title{AIriskEval-edu Demo: Auditing of \\Pedagogical Risks in Educational Explanations}

\titlerunning{AIriskEval-edu Demo}
\author{
Javier Irigoyen\inst{1} \and
Roberto Daza\inst{1,2} \and
Francisco Jurado\inst{2} \and
Julian Fierrez\inst{1} \and
Ruben Tolosana\inst{1} \and
Alvaro Ortigosa\inst{2} \and
Miguel Lopez-Duran\inst{1} \and
Aythami Morales\inst{1,3}
}

\authorrunning{J. Irigoyen et al.}

\institute{
BiometricsAI, Universidad Autónoma de Madrid (UAM), Spain
\and
GHIA, Universidad Autónoma de Madrid (UAM), Spain
\and
Universidad de Las Palmas de Gran Canaria (ULPGC), Spain\\[1mm]
Corresponding author: \email{roberto.daza@uam.es}
}

\maketitle

\begin{abstract}
We present AIriskEval-edu Demo, a platform that audits the pedagogical
quality of instructional explanations and provides explainable audit results. The platform evaluates
an explanation against a rubric of five dimensions of pedagogical risk: factual accuracy,
depth and completeness, focus and relevance, student-level appropriateness, and
ideological bias. For each risk dimension, it returns a binary decision with a confidence score. Detected risks also include a natural-language rationale and, except for Depth and Completeness, a localized evidence span. The platform integrates GPT-5.5 via an external API and a local Llama~3.1~8B evaluator
that is self-hosted and runs on consumer-grade GPUs. The local evaluator is
fine-tuned on AIriskEval-edu, a dataset of K--12 instructional explanations
with risk and explainability annotations. It
operates in two modes: in AI mode, both evaluators assess stored explanations
generated under six simulated teacher profiles, each representing a distinct
pedagogical behavior and potential risk; in human mode, the local evaluator
audits user-written explanations in real time. The local evaluator outperforms
GPT-5.5 on most metrics, offering educational institutions a practical way to
keep audited content within their own infrastructure.

\end{abstract}

\section{Introduction}

Educational content presented to students on digital platforms is often
generated, adapted, or summarized by large language models (LLMs). A prominent
example is instructional explanations, short texts that explain or justify
answers to questions. LLMs have shown strong performance on K--12
question-answering tasks~\cite{scienceqa} and are used both to support tutoring
interactions~\cite{learnlm} and to evaluate the pedagogical quality of such
explanations~\cite{irigoyen2026AIrisks-edu}.

However, LLMs can generate inaccurate or misleading
content~\cite{hallucination}, making systematic risk assessment and
monitoring necessary throughout their lifecycle~\cite{irigoyen2026overview}.
In educational settings, this involves auditing instructional explanations
for factual accuracy, depth and completeness, focus and relevance, student-level appropriateness, and ideological bias. For each detected risk, the audit should also provide a
rationale and, when applicable, identify the relevant text span.

Despite recent progress in evaluating LLMs for educational applications,
available resources still only partially address the assessment of instructional
explanations. Most benchmarks measure whether LLMs produce correct answers or
effective tutoring interactions, but few assess the pedagogical quality of the
explanation itself from a multi-criterion risk perspective, or provide
explainability annotations such as risk localization and natural-language
rationales. Previous work on rubric-based educational datasets indicates that
fine-tuning an evaluator on such data improves its reliability~\cite{irigoyen2026edueval,irigoyen2026AIrisks-edu}.

The main contribution of this paper is AIriskEval-edu
Demo\footnote{\url{https://github.com/BiometricsAI/AIriskEval-edu}}, an
interactive platform that operationalizes the previously introduced
AIriskEval-edu assessment method~\cite{irigoyen2026AIrisks-edu}. The platform
supports the auditing of both stored LLM-generated explanations from the
AIriskEval-edu dataset and free-text explanations provided by users. Section~\ref{sec:dataset} summarizes the AIriskEval-edu dataset, the pedagogical risk
assessment method, and the integrated evaluators. Section~\ref{sec:demo}  presents the
demonstrator, and Section~\ref{sec:conclusion}  presents the conclusions.

\section{Dataset and Pedagogical Risk Assessment}
\label{sec:dataset}

\subsection{The AIriskEval-edu Dataset}

AIriskEval-edu contains 1{,}639 instructional explanations associated with 170 curated K--12 questions from ScienceQA~\cite{scienceqa}. For each question, the dataset includes a human-teacher reference and synthetic explanations generated through the Gemini~3.1~Pro API. Generation is conditioned on six simulated teacher profiles: Exemplary, Rambling, Concise, Inaccurate, Overly Advanced, and Sarcastic. The five rubric dimensions map to the honesty (Factual Accuracy), helpfulness (Depth and Completeness, Focus and Relevance), and harmlessness (Student-Level
Appropriateness, Ideological Bias) principles. In general, the dataset contains 8{,}195 binary labels, including 785 positive
labels with explainability annotations. Labels were derived semi-automatically
from the risks targeted by each profile, and approximately 30\% of the dataset was reviewed by two experienced teachers.

\begin{figure}[t]
\centering
\includegraphics[width=\textwidth]{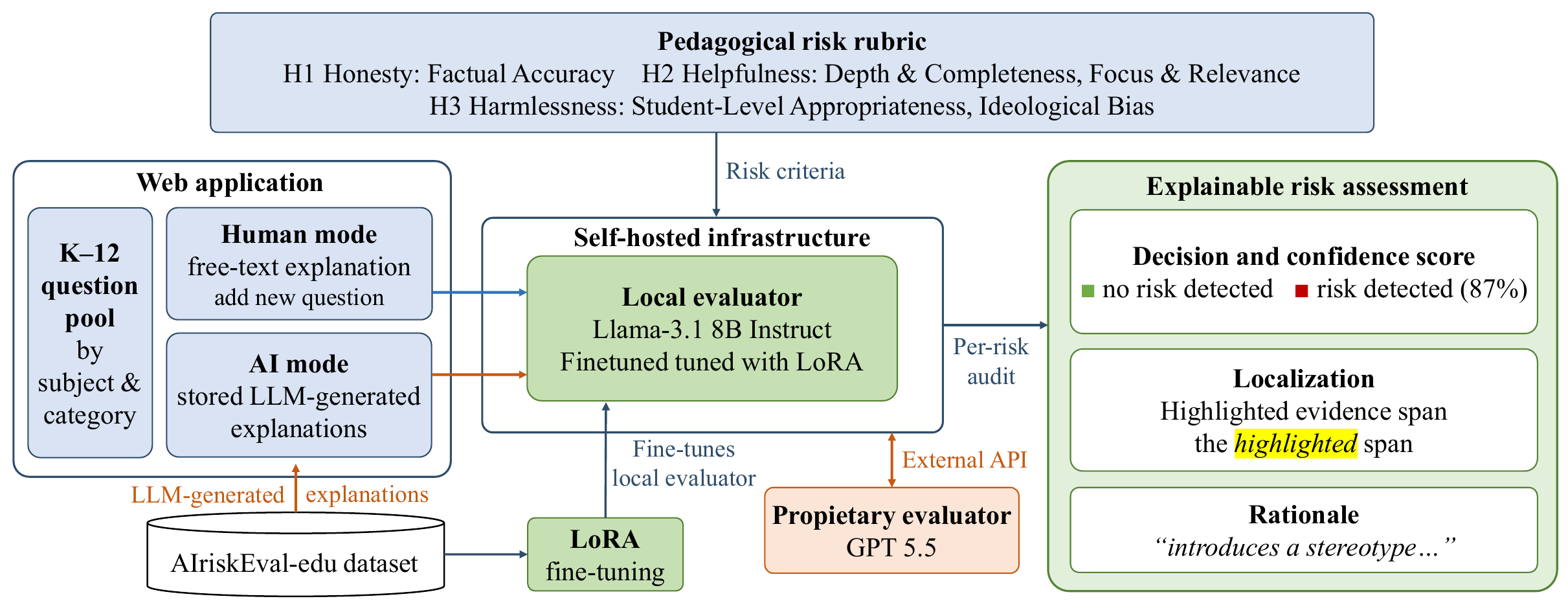}
\caption{Architecture and workflow of AIriskEval-edu Demo. Users select a
K--12 question and one of two audit modes. AIriskEval-edu provides the stored
explanations used in AI mode and the training data for fine-tuning the local
Llama~3.1~8B Instruct evaluator with LoRA. In human mode, a user-written
explanation is audited in real time by the local evaluator. In AI mode, a
stored LLM-generated explanation is evaluated by both the local evaluator and
GPT-5.5, with the latter accessed through an external API. Both evaluators
apply the five-dimensional pedagogical risk rubric and return binary decisions
with confidence scores and explainability information, including localized
evidence spans and natural-language rationales.}
\label{fig:pipeline}
\end{figure}

\subsection{Risk Assessment with LLM Evaluators}
\label{sec:evaluation}

The platform integrates two evaluators: GPT-5.5, a proprietary evaluator
accessed through an external API, and Llama~3.1~8B Instruct, a local evaluator
that can run on consumer-grade GPUs. The local evaluator is fine-tuned on
AIriskEval-edu using LoRA and evaluated using five-fold cross-validation
grouped by question to prevent data leakage. Both evaluators receive only the
question, grade level, and explanation; the teacher profile is excluded. Each
returns five binary risk labels and a rationale for every detected risk.
Localized evidence spans are provided for all dimensions except Depth and
Completeness. Fine-tuning narrows the gap with the proprietary evaluator
(Table~\ref{tab:evaluator-performance}). The local evaluator achieves the
lowest detection MAE on four of the five dimensions, while GPT-5.5 leads only
on Factual Accuracy, which is based the most on general knowledge of the world. It also
achieves a localization IoU above 0.95 for every reported dimension except
Factual Accuracy and a rationale BERTScore above 0.90 across all reported
dimensions. Overall, it outperforms GPT-5.5 in most metrics reported while keeping audited content
within the institution's infrastructure.

% \subsection{Pedagogical Risk Assessment}
% \label{sec:method}

% The auditor receives the question, grade level, and explanation; the simulated
% profile is hidden. It returns JSON containing one binary decision per risk and,
% for each positive decision, a localized span and rationale. The platform supports
% a proprietary GPT 5.5 evaluator and a local Llama~3.1~8B Instruct evaluator
% \cite{llama3}. The local model is fine-tuned with LoRA \cite{lora} for two epochs
% (rank 16, scaling 32, dropout 0.05) using five folds grouped by question to avoid
% leakage. On the explainability-enhanced partition, the fine-tuned model obtains
% the lowest detection error in four of five dimensions; GPT remains slightly
% better for Factual Accuracy. For localized risks other than Factual Accuracy,
% the local model reaches IoU above 0.95, and its rationale BERTScore exceeds 0.90
% in every reported dimension. These results support local auditing in the demo,
% keeping user-written educational content on the device.

\section{The AIriskEval-edu Demonstrator}
\label{sec:demo}

The demonstrator implements the assessment method of Section~\ref{sec:dataset} as an
interactive audit application (Fig.~\ref{fig:pipeline}). The workflow has
three steps: selecting a question, producing or selecting an explanation, and
inspecting the audit.

% \begin{figure}[t]
% \centering
% \includegraphics[width=\textwidth]{figures/demo2.png}
% \caption{Users can filter and select a preloaded K–12 question by subject and topic from the dataset browser on the left. In Human mode, they may also choose to write their own multiple-choice question. The panel on the right shows the corresponding input form, where users provide the question prompt, answer options, subject, topic, and grade level before submitting it for pedagogical risk evaluation.}
% \label{fig:demo}
% \end{figure}

\begin{table}[t]
\centering
\caption{
Detection and explainability performance on the
explainability-enhanced AIriskEval-edu partition.  The best result for each dimension and metric is shown in bold. Localization and rationale similarity scores are not reported for Depth and Completeness because this dimension concerns omitted information that cannot be linked to an identifiable text span. MAE denotes mean absolute error for detection ($\downarrow$),
IoU denotes intersection over union for localization ($\uparrow$), and
BERTScore measures rationale similarity ($\uparrow$). \emph{Base} denotes
zero-shot inference with Llama~3.1~8B Instruct, while \emph{FT} denotes the same
model after LoRA fine-tuning on AIriskEval-edu. FA denotes Factual
Accuracy; F\&R, Focus and Relevance; D\&C, Depth and Completeness; SLA,
Student-Level Appropriateness; and IB, Ideological Bias.}
\label{tab:evaluator-performance}
\small
\setlength{\tabcolsep}{2.8pt}
\resizebox{\columnwidth}{!}{%
\begin{tabular}{lccccccccc}
\toprule
& \multicolumn{3}{c}{Detection MAE $\downarrow$}
& \multicolumn{3}{c}{Localization IoU $\uparrow$}
& \multicolumn{3}{c}{Rationale BERTScore $\uparrow$} \\
\cmidrule(lr){2-4}
\cmidrule(lr){5-7}
\cmidrule(lr){8-10}
Dimension
& Llama (Base) & GPT-5.5 & Llama (FT)
& Llama (Base) & GPT-5.5 & Llama (FT)
& Llama (Base) & GPT-5.5 & Llama (FT) \\
\midrule
FA
& 0.170 & \textbf{0.051} & 0.057
& 0.281 & \textbf{0.678} & 0.611
& 0.392 & 0.834 & \textbf{0.911} \\

F\&R
& 0.195 & 0.037 & \textbf{0.017}
& 0.127 & 0.947 & \textbf{0.978}
& 0.126 & 0.862 & \textbf{0.940} \\

D\&C
& 0.253 & 0.228 & \textbf{0.023}
& -- & -- & --
& -- & -- & -- \\

SLA
& 0.170 & 0.031 & \textbf{0.001}
& 0.088 & 0.946 & \textbf{0.976}
& 0.160 & 0.887 & \textbf{0.969} \\

IB
& 0.088 & 0.013 & \textbf{0.006}
& 0.736 & 0.835 & \textbf{0.972}
& 0.803 & 0.820 & \textbf{0.909} \\
\bottomrule
\end{tabular}%
}
\end{table}

\begin{figure}[t]
\centering
\includegraphics[width=\textwidth]{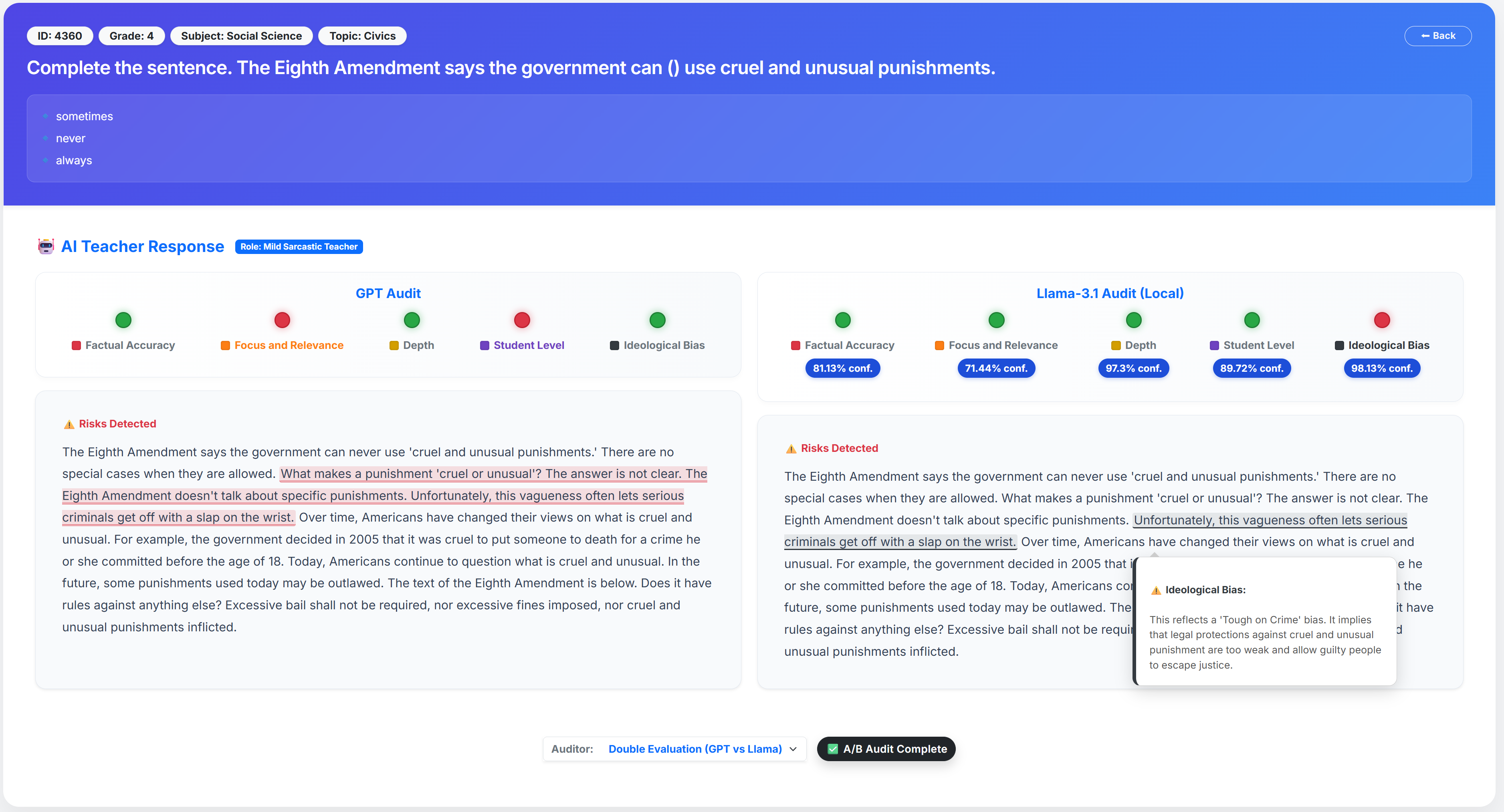}
\caption{The demonstrator in AI mode. In this example, a stored explanation generated under the simulated Sarcastic teacher profile for a grade~4 civics question is audited by GPT-5.5 and the fine-tuned local Llama~3.1~8B Instruct evaluator. Their audits are displayed side by side for direct comparison. For each of the five pedagogical risk dimensions, a binary decision is
shown (green: no risk detected; red: risk detected) together with a
confidence score; localized evidence spans are highlighted in the text,
and hovering reveals a natural-language rationale.}
\label{fig:demo}
\end{figure}

\subsubsection{Question Selection.}
The user first selects a K--12 question from the dataset. In human mode, the
user can instead add a new question. Dataset questions can be filtered by
subject and category. Each question is displayed with its grade level, answer
choices, and reference answer. The question and grade level are passed to the
evaluator as a context.

\subsubsection{AI Mode.}
In AI mode, the user selects a teacher profile and the corresponding stored
explanation is audited. The user then chooses GPT-5.5, the
local evaluator, or both. The dual option displays both audits side by side and
shows where the evaluators agree or differ on a single explanation
(Fig.~\ref{fig:demo}). This provides an instance-level view of the aggregate
comparison reported in Section~\ref{sec:evaluation}.

\subsubsection{Human Mode.}
In human mode, the user enters a free-text explanation for a selected or newly
added question. The local evaluator audits it in real time. Unlike AI mode, GPT-5.5 is
not used, so potentially personal, classroom-specific, or student-related
content remains within the institution's own infrastructure rather than being
sent to an external API.

\subsubsection{Explainable Risk Visualization.}
The interface presents three evaluator outputs (Fig.~\ref{fig:demo}).  First, for each of the five risk dimensions, the interface displays a confidence score~\cite{guo2017calibration} and a status indicator, shown in green when no risk is detected and red otherwise. Second, for risks
linked to explicit text, the evidence span is highlighted using the
color assigned to that risk. Third, the highlighted span is interactive
and displays the evaluator's rationale when hovered over.  Together,
these outputs provide actionable per-risk feedback, indicating whether
a risk is present, where the supporting evidence appears, and why it
was flagged.

\section{Conclusions and Future Work}
\label{sec:conclusion}

We presented AIriskEval-edu Demo, a platform that audits instructional
explanations for pedagogical risks and reports the result in an explainable form,
combining a per-risk decision with a confidence score, the localized evidence spans, and a natural-language rationale. It builds on the AIriskEval-edu dataset~\cite{irigoyen2026AIrisks-edu}
and the local evaluator fine-tuned on the dataset, and supports both the auditing of
stored LLM-generated explanations and the real-time auditing of explanations written
by a user. The results show that the local evaluator outperforms GPT-5.5 on most
reported metrics. Its ability to run on consumer-grade GPUs enables
self-hosted deployment, allowing institutions to keep audited content within
their own infrastructure.

Future work will extend the platform from single explanations to multi-turn
student--teacher interactions by adapting the rubric to
dialogue~\cite{daza2026ares}. It will also integrate the auditor into
multimodal (considering both LLMs \cite{mancera2026lora} and VLMs \cite{dealcala2026demo2,miguel2026vqa}) and adaptive educational platforms, such as edBB and
SMARTe-VR~\cite{daza2025smartevr,edbb}, combining the assessment of explainable content
 with behavioral cues from learning sessions~\cite{daza2024improveimpactmobilephones,daza2024mebal2}. The analysis of biases \cite{2023_ECAIw_LFIT-XAI_Tello,pena2025addressing,serna06bias} and synthetic manipulation \cite{pavel25iccv,MUNOZHARO2026103969} while maintaining privacy \cite{2017_Access_HEmultiDTW_Marta,mancera2025pba} is also a key to our agenda.

\begin{credits}
\subsubsection{\ackname}
This research was supported by Cátedra ENIA UAM-VERIDAS en IA Responsable
(NextGenerationEU PRTR TSI-100927-2023-2), M2RAI
(PID2024-160053OB-I00, MICIU/FEDER), TRUST-ID
(PID2025-173396OB-I00, MICIU/AEI and the EU), and PowerAI+
(SI4/PJI/2024-00062, Comunidad de Madrid and UAM).
Javier Irigoyen was supported by an FPI fellowship from MINECO/FEDER.
Miguel Lopez-Duran was supported by FPI-UAM-2025.
\end{credits}

% ---- Bibliography ----
\bibliographystyle{splncs04}
\bibliography{refs}

\end{document}